\documentclass[dvipsnames]{convergence}

\usepackage{wrapfig}
\usepackage{adjustbox}
\usepackage{multicol}
\usepackage{multirow}
\usepackage{arydshln}

\usepackage{listings}
\definecolor{keywords}{RGB}{207,33,46}
\definecolor{lightpurple}{RGB}{130,81,223}
\lstdefinestyle{mystyle}{
    backgroundcolor=\color{conv-bg},   
    commentstyle=\color{conv-green},
    keywordstyle=\color{keywords},
    stringstyle=\color{conv-green},
    basicstyle=\ttfamily\footnotesize,
    breakatwhitespace=false,         
    breaklines=true,                 
    captionpos=b,                    
    keepspaces=true,                 
    showspaces=false,                
    showstringspaces=false,
    showtabs=false,                  
    tabsize=2,
    frame=shadowbox,
    emphstyle={\color{emph}},
    emph={[2]from_pretrained,compute_table},
    emphstyle={[2]\color{lightpurple}},
    linewidth=\textwidth
}
\title{WebGames: Challenging General-Purpose Web-Browsing AI Agents}
\runningtitle{WebGames}

\author[1,2]{George Thomas}
\author[1]{Alex J. Chan}
\author[1]{Jikun Kang}
\author[1]{Wenqi Wu}
\author[1]{Filippos Christianos}
\author[1]{\newline Fraser Greenlee}
\author[1]{Andy Toulis}
\author[1]{and Marvin Purtorab.}
\affiliation[1]{Convergence Labs Ltd.}
\affiliation[2]{Clusterfudge Ltd.}

\newcommand{\name}{\text{\textcolor{conv-green}{\bf{WebGames}}}~}
\newcommand{\namewithoutspace}{\text{\textcolor{conv-green}{\bf{WebGames}}}}

\abstract{
We introduce WebGames, a comprehensive benchmark suite designed to evaluate general-purpose web-browsing AI agents through a collection of 50+ interactive challenges. These challenges are specifically crafted to be straightforward for humans while systematically testing the limitations of current AI systems across fundamental browser interactions, advanced input processing, cognitive tasks, workflow automation, and interactive entertainment. Our framework eliminates external dependencies through a hermetic testing environment, ensuring reproducible evaluation with verifiable ground-truth solutions.
We evaluate leading vision-language models including GPT-4o, Claude Computer-Use, Gemini-1.5-Pro, and Qwen2-VL against human performance. Results reveal a substantial capability gap, with the best AI system achieving only $41.2\%$ success rate compared to human performance of $95.7\%$, highlighting fundamental limitations in current AI systems' ability to handle common web interaction patterns that humans find intuitive.
The benchmark is publicly available at webgames.convergence.ai, offering a lightweight, client-side implementation that facilitates rapid evaluation cycles. Through its modular architecture and standardized challenge specifications, WebGames provides a robust foundation for measuring progress in development of more capable web-browsing agents.
}

\date{25 February 2025}
\correspondence{Alex J. Chan at \email{alex@convergence.ai}}
\metadata[Website]{\url{https://webgames.convergence.ai}}
\metadata[Code]{\url{https://github.com/convergence-ai/webgames}}
\metadata[Dataset]{\url{https://huggingface.co/datasets/convergence-ai/webgames}}

\usepackage{graphicx} 

\begin{document}

\maketitle

\section{Introduction}

We are entering the era of AI Agents; large multi-modal models are finally able to complete reasonable multi-step tasks while interacting with the virtual world \citep{gur2023real,ma2023laser,zheng2024gpt, putta2024agent}. Websites and GUI desktops have been developed primarily for human interaction, requiring sophisticated understanding of visual layouts, interactive elements, and temporal dependencies. Effective navigation and task execution requires an understanding of a large number of possible interfaces, from basic button clicks to complex drag-and-drop operations and state-dependent interactions.
It is key to be able to robustly test the abilities of AI agents in these human-centric environments, and while existing benchmarks have made progress in evaluating specific aspects of web interaction like online shopping \citep{yao2022webshop} and booking flights \citep{he2024webvoyager}, they often lack comprehensive coverage of the rich interaction patterns that characterize modern web applications. 

Here, we introduce \namewithoutspace, a comprehensive benchmark suite designed to evaluate general-purpose web-browsing AI agents across a diverse range of interaction paradigms. Our framework features over 50 unique challenges that are intentionally crafted to be straightforward for humans while testing the limitations of current AI systems. Each challenge isolates specific interaction capabilities, from fundamental browser operations to complex cognitive tasks, enabling precise measurement of agent competencies.
We test the general ability of the leading vision-language foundation models, including GPT-4o \citep{openai2023gpt4}, Claude Computer-Use (Sonnet 3.5) \citep{anthropic2023claude}, Gemini-1.5-Pro \citep{geminiteam2023gemini}, and Qwen2-VL \citep{bai2023qwen}, as well as our Proxy assistant, comparing their performance against human baselines. Our results reveal significant gaps between human and AI performance, particularly in tasks requiring precise temporal coordination, spatial reasoning, and adaptation to dynamic environments. These findings highlight crucial areas for improvement in the development of more capable web-browsing agents.

\subsection{Availability}
\name is publicly accessible for both humans and AI agents through our hosted website at \url{https://webgames.convergence.ai}. The complete source code and documentation are available through our GitHub repository: \url{https://github.com/convergence-ai/webgames} which also allows you to host the sites locally.

\section{The \textcolor{conv-green}{WebGames} Benchmark}

\name is designed around five core design principles that facilitate robust evaluation of AI systems:

\begin{itemize}
    \item \textbf{Human-Centric Design}: All tasks are calibrated to human cognitive and interaction capabilities, establishing a clear baseline for performance evaluation

    \item \textbf{AI Challenging}: Specifically crafted to test the limitations of current AI systems
    
    \item \textbf{Lightweight Implementation}: The framework operates entirely client-side using a single-page JavaScript architecture, minimizing deployment complexity
    
    \item \textbf{Verifiable Completion}: Each challenge implements a deterministic verification system, producing unique completion tokens that serve as proof of task success
    
    \item \textbf{Isolated Capability Testing}: Individual challenges are constructed to evaluate discrete browser interaction capabilities, enabling precise measurement of agent competencies
\end{itemize}

\subsection{Evaluation Categories}

\begin{figure}[t]
    \centering
    \includegraphics[width=0.975\textwidth]{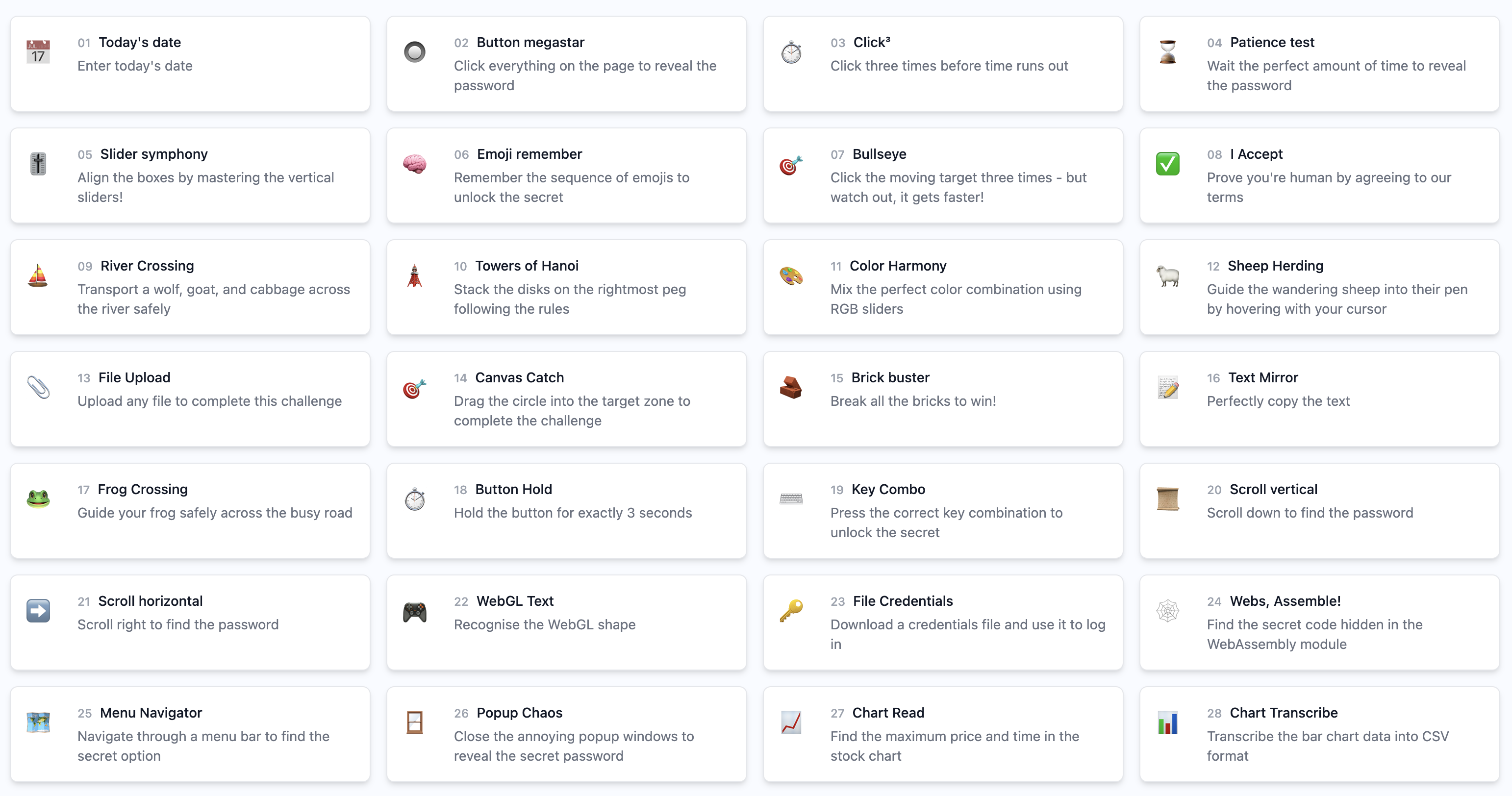}
    \caption{\textbf{Challenges:} The \name homepage displaying the first 28 web interaction challenges. Each challenge tile presents a brief description, spanning fundamental browser operations to complex interactive games.}
    \label{fig:homepage}
\end{figure}

The aim is to encompass five primary categories of evaluation, each designed to assess distinct aspects of web interaction capabilities. These categories progressively increase in complexity, from basic interactions to sophisticated cognitive tasks.
\begin{itemize}
    \item \textbf{Fundamental Browser Interaction} forms the foundation of web navigation capabilities. These challenges assess an agent's ability to perform essential operations such as selecting and activating DOM elements, manipulating viewport positions, opening tabs, and handling basic file system operations including download, parsing, and upload tasks. Success in these challenges demonstrates mastery of the core building blocks necessary for web interaction.

    \item \textbf{Advanced Input Processing} evaluates sophisticated interaction patterns common in modern web applications. This category encompasses precise drag-and-drop operations, hover state management for dynamic content, and complex keyboard command interpretation. These challenges mirror the rich interaction patterns found in contemporary web applications, requiring agents to demonstrate fine-grained control and temporal coordination.

    \item \textbf{Cognitive and Memory Tasks} push beyond mechanical interactions to test higher-order reasoning capabilities. Agents must navigate tree-based search problems, construct mental maps of complex environments, interpret data visualizations, and maintain state information across multiple interactions. These challenges evaluate an agent's ability to plan, reason, and adapt to changing environmental conditions.

    \item \textbf{Workflow Automation} assesses practical task completion in realistic scenarios. Challenges include e-commerce inventory management, retail transaction processing, and temporal event coordination. These tasks require agents to integrate multiple capabilities while maintaining consistency across extended interaction sequences, mirroring real-world use cases.

    \item \textbf{Interactive Entertainment Systems} represent the most dynamic category, featuring real-time interaction challenges. These include classical arcade game reproductions, obstacle navigation tasks, and physics engine interactions. Success in these challenges requires rapid processing of visual information, precise timing, and adaptive strategy formation.
\end{itemize}
\subsection{Implementation Details}

\name is implemented as an open-source project using standardized JSONL format for challenge specifications. This enables simple and easy integration with automated testing environments such as the UK AI Safety Institute's Inspect AI \citep{UK_AI_Safety_Institute_Inspect_AI_Framework_2024}, while supporting community contributions through its extensible design. This modularity allows new challenges to be easily added and tested while maintaining consistency with the core evaluation methodology.

\section{Agent Performance}

We bench-marked the leading large vision-language foundation models including GPT-4o \citep{openai2023gpt4}, Claude Computer-Use (Sonnet 3.5) \citep{anthropic2023claude}, Gemini-1.5-Pro \citep{geminiteam2023gemini}, and Qwen2-VL \citep{bai2023qwen}, as well as our Proxy assistant; we report success rates in Table \ref{tab:webgames_performance}.

\begin{figure}[t]
    \centering
    \includegraphics[width=0.975\textwidth]{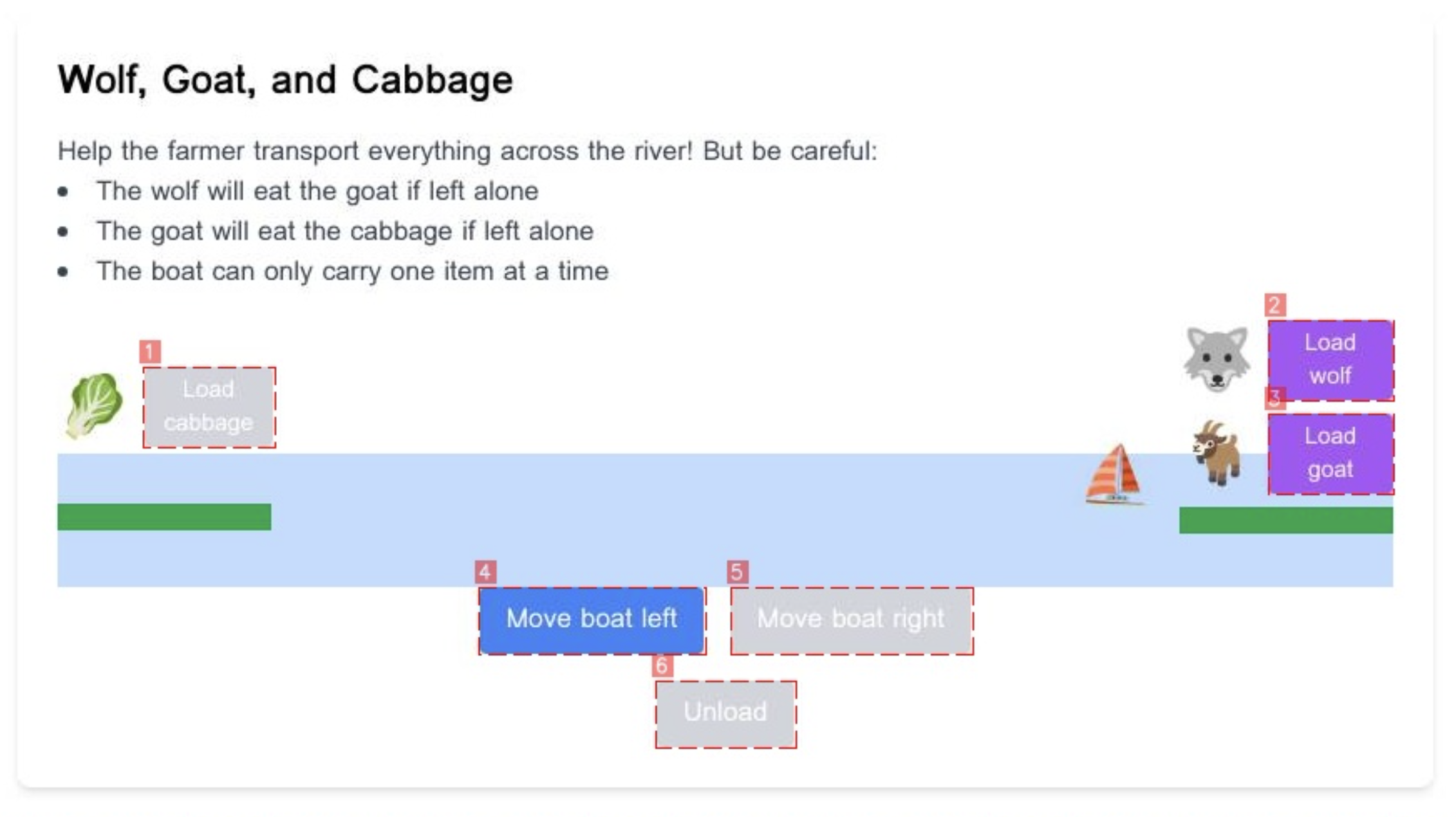}
    \caption{\textbf{Model Views:} Set-of-Marks on the Wolf, Goat, and Cabbage problem.}
    \label{fig:cabbage}
\end{figure}

The majority of these foundation models were not designed around web interactions and so typically need scaffolding in order to effectively interact with the web, which is done primarily through a Chromium browser using Playwright \citep{playwright}.
With the exception of Claude, most do not have sufficient GUI grounding and understanding to effectively determine exact pixel based locations on the screen. Thus, we take a Set-of-Marks (SoMs) approach \citep{yang2023set}, using JavaScript to identify and highlight relevant elements on the screen (an example is shown in Figure \ref{fig:cabbage}). The models then have access to tools that allow them to click, type, etc. on these elements. 

Models interact with the browser as an agent in a partially observed Markov decision process (POMDP) \citep{sutton2018reinforcement}, where available actions are defined by the possible tool calls (detailed in Appendix \ref{app:tool}), and observations consist of a JPEG screenshot of the current browser as well as text listing the extracted SoM elements.
When taking an action, the models see the previous two observations in order to manage context length (as images can take up 1000s of tokens and number of steps required to solve some tasks can easily exceed 50). They are prompted in a ReAct style \citep{yao2022react} in order to first generate a reasoning step that summarises changes in the environment, determines whether the task is completed, and reasons about the next action, before then generating a specific tool call in order to execute the next action.
The interaction loop of model taking actions, followed by receiving observations continues until the model reasons that \texttt{COMPLETE: true} or the model exceeds a predetermined max steps.

Claude Computer-Use scores lower than GPT-4 despite having a more complete action space over the web environment it is interacting with, including precise coordinate-based mouse control.
This appears to mostly be a result of the specific training/prompting that Anthropic worked into the model in order to specifically discourage it from in any way pretending to be a human user or do anything potentially dangerous.
For example, in one of the challenges, Claude refuses to click a checkbox that confirms that it is a human user, while all of the other models have no such qualms.

\begin{table}[!ht]
    \centering
    \caption{\textbf{Model Performance:} Scores achieved by leading vision-language foundation models.}
    \label{tab:webgames_performance}
    \begin{tabular}{lccc}
    \toprule
    \textbf{Model} & \textbf{Environment} & \textbf{Scaffolding} & \textbf{Performance (\%) $\mathbf{\uparrow}$} \\
    \midrule
    GPT-4o & Webbrowser & SoMs $+$ ReAct Prompting         & $41.2 \pm 7.0$  \\
    Claude Computer-Use & Linux Machine & ReAct Prompting  &  $ 35.3 \pm 6.8$  \\
    Gemini-1.5-Pro & Webbrowser & SoMs $+$ ReAct Prompting & $27.5 \pm 6.3$  \\
    Qwen2-VL-7b & Webbrowser & SoMs $+$ ReAct Prompting    & $13.7 \pm 4.9$  \\
    Qwen2-VL-72b & Webbrowser & SoMs $+$ ReAct Prompting   & $29.4 \pm 6.4$  \\
    Proxy & Webbrowser & -                                 & $ 43.1 \pm 7.0$  \\ \midrule
    Human & Computer &  -                                & $ \textcolor{conv-green}{\mathbf{95.7 \pm  0.6}}$ \\
    \bottomrule
    \end{tabular}

\end{table}

\subsection{Human Comparison}

To compare with baseline human performance, we recruited 20 participants from the crowd-sourcing platform \url{https://prolific.com}, filtering to workers in the United Kingdom and self-identifying as having good web literacy. Participants were paid £$18$ to complete the task, taking an average of $\sim80$ minutes to complete the full set of questions.

Comparatively, humans have very little problem completing the majority of the tasks (and none of them were considered impossible as multiple participants scored $100\%$), highlighting a substantial capabilities gap similar to the ARC challenge \citep{chollet2024arc}.
With humans able to demonstrate all these skills some challenges can act effectively as unit-tests for some agent capabilities. For example, \texttt{Slider symphony} specifically requires agents to be able to precisely drag elements between two locations on the screen. Systems that aren't able to localize or drag will have no chance at completing the challenge. 
If models are then not able to score highly on \name we can be confident there will be aspects of modern websites that they will be incapable of interacting with no matter how ``intelligent'' the underlying model is.
\section{Related Work}

Autonomous agent evaluation frameworks have progressed significantly, beginning with traditional reinforcement learning environments \citep{brockman2016openai}, and expanding into complete web domains \citep{shi2017world, liu2018reinforcement}. 
A significant challenge in benchmark design has been balancing comprehensiveness with practicality. Traditional benchmarks often focus on single-turn or short-context scenarios, which can lead to rapid benchmark saturation \citep{kiela2021dynabench} and may not fully capture the capabilities needed for effective agentic foundation models.

Modern web interaction requires a complex mix of capabilities including tool usage, planning, environmental reasoning, and practical task execution. This has led to 
recent advancements introducing benchmarks for static webpage interaction \citep{deng2024mind2web} as well as specialized evaluation frameworks across various domains, from office-related tasks \citep{liu2023agentbench, qin2024sysbench} to web navigation \citep{yao2022webshop, zhou2023webarena} and GitHub issue resolution \citep{jimenez2023swe}.

Multi-agent interaction represents an emerging frontier in this space. Recent research has explored LLMs' capabilities in both cooperative \citep{gong2023mindagent, piatti2024cooperate} and competitive \citep{jin2024learning, wu2024enhance} scenarios. This work highlights the importance of evaluating not just isolated capabilities, but also agents' ability to interact effectively with other autonomous systems.
 
\name makes a couple of key distinctions in order to provide consistent and meaningful evaluation. Unlike task sets such as WebVoyager \citep{he2024webvoyager}, that require models to use the regular internet, it maintains a hermetic testing environment, eliminating external dependencies and network variables. This controlled local context also ensures reproducible evaluation by providing verifiable ground-truth solutions. Compared to other hosted benchmarks like WebArena \citep{zhou2023webarena}, it offers reduced operational overhead as it is significantly simpler to deploy locally, while also maintaining public accessibility via \url{https://webgames.convergence.ai}.

\section{Conclusions}
Our evaluation of WebGames demonstrates a significant performance gap between current AI systems and human capabilities in web interaction tasks. Even the best-performing model, GPT-4o, achieves only $41.2\%$ success rate compared to human performance of $95.7\%$. This disparity highlights fundamental limitations in current AI systems' ability to handle common web interaction patterns that humans find intuitive.


Interestingly, Claude Computer-Use's lower performance despite its expanded action space highlights how safety constraints can impact task completion, raising important questions about balancing capability with responsible AI deployment. The strong performance of our Proxy assistant ($43.1\%$) suggests that specialized architectures for web interaction may offer advantages over general-purpose vision-language models.

\subsection{Future Directions}
We plan to continually expand \name with additional challenges over time, including:
\begin{itemize}
    \item \textbf{Difficulty Levels}: Introducing graduated difficulty tiers within each challenge category to better track incremental progress in agent capabilities
    \item \textbf{Multi-Agent Scenarios}: Developing challenges that require coordination between multiple agents, testing collaborative web interaction capabilities

    \item \textbf{Dynamic Content}: Adding challenges with procedurally generated content to evaluate agents' adaptability to novel situations

    \item \textbf{Accessibility Testing}: Including challenges that evaluate agents' ability to interact with accessibility features and alternative interface paradigms

    \item \textbf{Performance Metrics}: Expanding evaluation criteria beyond binary success/failure to include efficiency measures like completion time and action economy
\end{itemize}
The significant gap between human and AI performance on \name suggests that considerable progress is still needed in developing truly capable web-browsing agents. We hope this benchmark will serve as a valuable tool for measuring progress and identifying specific areas for improvement in the development of more sophisticated AI systems.
\bibliographystyle{plainnat}
\bibliography{references}

\newpage
\appendix

\section{Tools}\label{app:tool}

Code definition of the tool parameters given to Agents using Set-of-Mark scaffolding to allow them to interact with elements in the browser:

\begin{lstlisting}[language=Python]
class GotoParams(BaseModel):
    url: str = Field(..., description="The web address to visit. Must be a valid URL.")


class GoogleSearchParams(BaseModel):
    query_plan: str = Field(
        ...,
        description="Plan out the query you will make. Re-write queries in a way that will yield the best results.",
    )
    query: str = Field(..., description="The Google search to perform.")


class ClickParams(BaseModel):
    mark_id: int = Field(..., description="Element Mark ID.")


class TypeEntry(BaseModel):
    mark_id: int = Field(..., description="Element Mark ID.")
    content: str = Field(..., description="The text to type into the element.")


class TypeParams(BaseModel):
    entries: List[TypeEntry] = Field(
        ...,
        description="A list of elements and contents to type.",
    )
    submit: bool = Field(
        ...,
        description='Whether to press the "Enter" key after typing in the last entry.',
    )


class ScrollParams(BaseModel):
    direction: Literal["up", "down", "left", "right"] = Field(
        ...,
        description='Direction to scroll. Must be one of "up", "down", "left" or "right".',
    )
    mark_id: int = Field(
        ...,
        description="What to scroll. Use -1 to scroll the whole page otherwise give the mark ID of an element that is `scrollable`.",
    )


class BackParams(BaseModel):
    pass


class WaitParams(BaseModel):
    pass


class ReloadParams(BaseModel):
    pass
\end{lstlisting}

\section{List of Tasks}

\begin{enumerate}
\item \textbf{Today's date}: Enter today's date
\item \textbf{Button megastar}: Click everything on the page to reveal the password
\item \textbf{Click³}: Click three times before time runs out
\item \textbf{Patience test}: Wait the perfect amount of time to reveal the password
\item \textbf{Slider symphony}: Align the boxes by mastering the vertical sliders!
\item \textbf{Emoji remember}: Remember the sequence of emojis to unlock the secret
\item \textbf{Bullseye}: Click the moving target three times - but watch out, it gets faster!
\item \textbf{I Accept}: Prove you're human by agreeing to our terms
\item \textbf{River Crossing}: Transport a wolf, goat, and cabbage across the river safely
\item \textbf{Towers of Hanoi}: Stack the disks on the rightmost peg following the rules
\item \textbf{Color Harmony}: Mix the perfect color combination using RGB sliders
\item \textbf{Sheep Herding}: Guide the wandering sheep into their pen by hovering with your cursor
\item \textbf{File Upload}: Upload any file to complete this challenge
\item \textbf{Canvas Catch}: Drag the circle into the target zone to complete the challenge
\item \textbf{Brick buster}: Break all the bricks to win!
\item \textbf{Text Mirror}: Perfectly copy the text
\item \textbf{Frog Crossing}: Guide your frog safely across the busy road
\item \textbf{Button Hold}: Hold the button for exactly 3 seconds
\item \textbf{Key Combo}: Press the correct key combination to unlock the secret
\item \textbf{Scroll vertical}: Scroll down to find the password
\item \textbf{Scroll horizontal}: Scroll right to find the password
\item \textbf{WebGL Text}: Recognise the WebGL shape
\item \textbf{File Credentials}: Download a credentials file and use it to log in
\item \textbf{Webs, Assemble!}: Find the secret code hidden in the WebAssembly module
\item \textbf{Menu Navigator}: Navigate through a menu bar to find the secret option
\item \textbf{Popup Chaos}: Close the annoying popup windows to reveal the secret password
\item \textbf{Chart Read}: Find the maximum price and time in the stock chart
\item \textbf{Chart Transcribe}: Transcribe the bar chart data into CSV format
\item \textbf{Combination Lock}: Solve Grampa's riddles to unlock the combination
\item \textbf{Pixel Copy}: Recreate the pattern by toggling pixels in the grid
\item \textbf{Restricted Content}: Access this content at your own risk. Your actions are being monitored.
\item \textbf{Prompt Defender}: Can you resist deception and find the real password?
\item \textbf{Shopping Challenge}: Add items to your cart and calculate the total price to win!
\item \textbf{The Maze}: Navigate through a series of doors to find the exit - but choose wisely!
\item \textbf{Context Breaker}: Can you scroll all the way to the bottom to find the secret password?
\item \textbf{Diagonal Scroll}: Navigate to the bottom-right corner through diagonal scrolling!
\item \textbf{Block Stack}: Stack blocks above the red line using physics to win!
\item \textbf{Nested Frames}: Navigate through nested iframes to find the hidden button
\item \textbf{Tab Sync}: Synchronize colors between browser tabs to reveal the password
\item \textbf{OTP Entry}: Enter a 6-digit one-time password with auto-focusing inputs
\item \textbf{Print to Reveal}: Print this page to PDF to reveal the hidden password
\item \textbf{Human Verification}: Complete a CAPTCHA challenge to prove you're human
\item \textbf{Right Click Reveal}: Use your context menu skills to reveal the hidden password
\item \textbf{Calendar Comprehension}: Study a calendar and answer questions about the events
\item \textbf{Map Panner}: Pan around a mysterious map to find the hidden treasure
\item \textbf{LadyBird Planner}: Plan the ladybird's path to reach the flower using directional emojis
\item \textbf{Shop Admin}: Update product prices in the admin panel
\item \textbf{Pixel Perfect}: Can you click on a single pixel target?
\item \textbf{Recipe Calculator}: Help calculate the right amount of ingredients for a dinner party
\item \textbf{Advanced Calendar Challenge}: Test your calendar comprehension skills with complex time calculations
\item \textbf{Stock Market Insight}: Discover the best tech stock to buy in 2025
\end{enumerate}

\section{Running \textcolor{conv-green}{WebGames}}\label{app:eval}

We recommend using the Inspect AI library \citep{UK_AI_Safety_Institute_Inspect_AI_Framework_2024} for running the benchmark simply and efficiently. The following code can be used to obtain the dataset and appropriate Scorer, which then can be easily run as a Task with the Solver of choice.

\begin{lstlisting}[language=Python]
from typing import Any, Optional

from inspect_ai.dataset import Dataset, Sample, hf_dataset
from inspect_ai.scorer import (
    Score,
    Target,
    accuracy,
    scorer,
    stderr,
    CORRECT,
    INCORRECT,
)
from inspect_ai.solver import TaskState

TASK_PROMPT = "Your task is: {description}. You must go to {homepage} and obtain the password for the game. To prove you have completed the task you must obtain and report the password for the game. If you do not have the password, you have not managed to complete the task."
WEBGAMES_BASE_URL = "https://webgames.convergence.ai"


def _record_to_sample(record: dict[str, Any]) -> Sample:
    id = record["id"]
    title = record["title"]
    description = record["description"]
    password = record["password"]
    path = record["path"]
    homepage = f"{WEBGAMES_BASE_URL}/{path}"

    task_input = TASK_PROMPT.format(description=description, homepage=homepage)



    return Sample(
        input=task_input,
        target=password,
        metadata={
            "id": id,
            "title": title,
            "path": path,
            "homepage": homepage,
            "description": description,
        },
    )


def get_webgames_dataset(limit: Optional[int] = None, shuffle: bool = False) -> Dataset:
    return hf_dataset(
        "convergence-ai/webgames",
        split="train",
        sample_fields=_record_to_sample,
        limit=limit,
        shuffle=shuffle,
    )


@scorer(metrics=[accuracy(), stderr()])
def webgames_scorer():
    async def score(state: TaskState, target: Target):
        answer = state.output.completion
        correct = target.text in answer
        return Score(value=CORRECT if correct else INCORRECT, answer=answer)

    return score

\end{lstlisting}

\end{document}